\documentclass{article} 
\usepackage[final,nonatbib]{nips_2016} 
\usepackage{hhline}
\usepackage{graphicx}
\usepackage{caption}
\usepackage{subcaption}
\usepackage[numbers]{natbib}
\usepackage [autostyle, english = american]{csquotes}
\MakeOuterQuote{"}
\title{Identifying Nominals with No Head Match Co-references Using Deep Learning}

\author{
Matthew Stone \\
\texttt{mdstone@stanford.edu} \\
\And
Ramnik Arora \\
\texttt{rarora@cs.stanford.edu} \\
Department of Computer Science \\
Stanford University
}

\begin{document}

\maketitle

\begin{abstract}
Identifying nominals with no head match is a long-standing challenge in coreference resolution with current systems performing significantly worse than humans. In this paper we present a new neural network architecture which outperforms the current state-of-the-art system on the English portion of the CoNLL 2012 Shared Task. This is done by using a logistic regression on features produced by two submodels, one of which is has the architecture proposed in \cite{DBLP:conf/emnlp/ClarkM16} while the other combines domain specific embeddings of the antecedent and the mention. We also propose some simple additional features which seem to improve performance for all models substantially, increasing $F_1$ by almost 4\% on basic logistic regression and other complex models.
\end{abstract}

\section{Introduction}
Coreference resolution is a domain of Natural Language Processing which is focused on finding all of the references to the same real world entity within a body of text. Applications of coreference resolution include full text understanding (for discourse understanding), machine translation (gender, numbers), text summarization, and question-answering systems.

Mentions, of which pairs can potentially form coreferences, come in three forms: named, pronominal or nominal. Named mentioned include a specific name, such as "President Obama." Pronominal mentions include the use of a pronoun but without naming the entity, such as "he" when referring to Obama. A nominal is the most vague and is neither named nor pronominal. A nominal mention could be "the President". Another nominal mention could be "the man guarded by the secret service." The first nominal example would have a head match with "President Obama" because of the overlap of the word "President". A nominal without a head match would be the second example.

This paper focuses on this match type using deep learning. One of the largest challenges with matching nominals with no head match is that a much deeper understanding of the mention pair is needed, including some modeling of the semantic meaning of each mention.


Traditional approaches to coreference began with Hobb's naive algorithm \cite{Hobbs:1986:RPR:21922.24343}. This approach embodies the semantic or structural methodology behind anaphora resolution through a rules-based system. This algorithm is one of the early baselines for coreference.

There are three major modeling approaches to coreference: 
\begin{itemize}
	\item \textbf{Mention pair models}, which is the type this paper will attempt to implement, make independent pair-wise decisions. Typically these models also implement a reconciliation mechanism as a final stage before evaluation. These mechanisms map links that are implied by the model predictions\cite{Hoste2016}.
	\item \textbf{Mention ranking models} are attempting to solve a ranking problem by listing all possible antecedents and optimizing a ranking system to find the best or most likely antecedent mention \cite{Han:2011:GEM:2002472.2002592}.
	\item \textbf{Entity mention models} attempt to link all mentions of one entity in a cluster model by selectively choosing which a clustering algorithm that reduces errors on large scale mention clusters \cite{DBLP:journals/corr/ClarkM16}.
\end{itemize}

Current state of the art models have a CoNLL $F_1$ \cite{pradhan-etal-conll-st-2012-ontonotes} score of almost 69\% \cite{DBLP:conf/emnlp/ClarkM16} in aggregate while only 18.9\% for corefences with no head match. Humans on the other hand score over 60\% on similar tasks. Nominals with no head match need to make much more use of context. In particular, they leverage the word vector embeddings for related semantic meaning to identify these types of coreferences. With this in mind, the approaches found in this paper will emphasize aspects of word vector alignment. We'll be using a mention-pair model treating all mentions-pairs as independent, without any post reconciliation implemented as yet. One of the advancements in \cite{DBLP:conf/emnlp/ClarkM16} was post reconciliation via cluster resolution which we did not replicate. If A and B, as well as B and C are identified as corefernces, then the subsequent step would evaluate A and C, and either cluster A and C or break one of the two other links.

In our model, we increase the test $F_1$ on a 2\% negative sample downsample from 0.49 for logistic regression baseline and 0.54 for state-of-the-art architecture in \cite{DBLP:conf/emnlp/ClarkM16} to 0.61 using a combination of new features and intelligent architecture. Note that these are not CoNLL $F_1$ scores, since this only applies to independent intra-document pairwise predictions rather than the downstream clustering task.

\section{Data \& Evaluation}
We use the data provided by the CoNLL 2012 Shared Task on English \cite{pradhan-etal-conll-st-2012-ontonotes}. 

\subsection{CoNLL 2012 Shared Task}
The CoNLL 2012 shared task dataset provides 2802 documents in training, 343 documents in dev and 348 documents in test. The mentions, and the co-reference pairs and types are tagged in this dataset. We filter on nominal mentions and co-reference pairs which have no head-match. We structure our dataset as an evaluation of direct pair matches, and build deep learning models directly around this subset. In this paper, we treat all mentions as independent and try to learn simply based on input features.

\begin{figure}[h]
	\begin{center}
		\includegraphics[width=\linewidth]{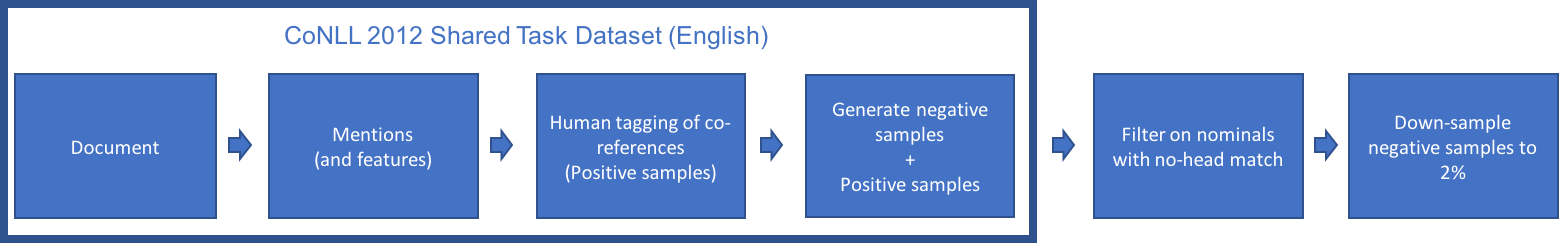}
	\end{center}
	\caption{Schematic diagram of the data}
	\label{data}
\end{figure}

The class label counts in the dataset are heavily skewed. If there are $N$ mentions in a documents, there will be $N^2$ total mention pairs and only a few of them will be co-references. This means that the dataset is heavily skewed with significantly more negative examples than positive examples. In our training data we have 7,643,471 negative samples while 8,010 positive samples. We've downsampled the negative samples to 2\% while maintaining 100\% of our positive samples. Even with this downsampling, only 5\% of our modified samples are positive mentions. See figure \ref{data} for schematic representation of data-generating process. 

\subsection{Evaluation Metrics and Nuances}
\begin{figure}[h]
	\begin{center}
		\includegraphics[width=0.5\linewidth]{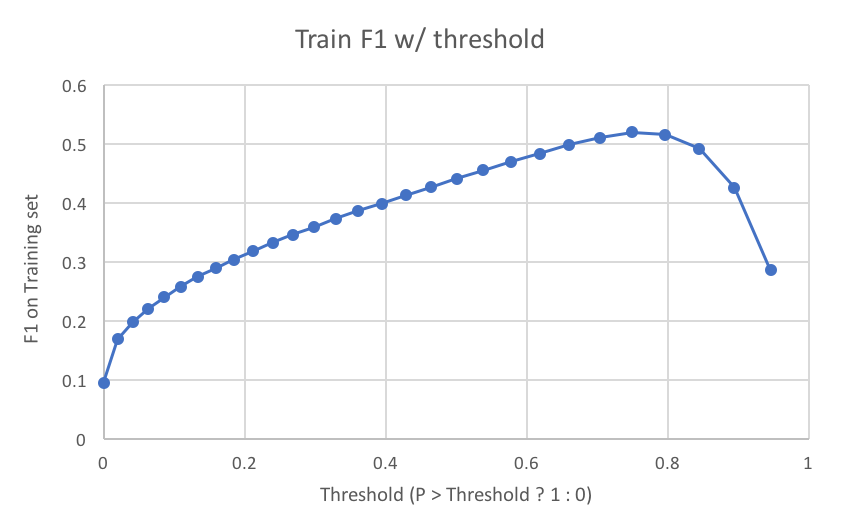}
	\end{center}
	\caption{$F_1$ is very sensitive to the threshold chosen}
	\label{f1_score}
\end{figure}

For co-reference classification, CoNLL $F_1$, which is an average of MUC \cite{Vilain:1995:MCS:1072399.1072405}, $B^3_1$ \cite{Bagga98algorithmsfor} and CEAF \cite{Luo:2005:CRP:1220575.1220579} $F_1$ scores, is common evaluation metric. While we've used $F_1$ as an evaluation metric, it has shortcomings. Since $F_1$ scores are based on a strict \{0, 1\} classification, while the model predicts a probability estimate, they are very sensitive to thresholds. This is evidenced by figure \ref{f1_score} where for the same probability outputs, we can have drastically varying $F_1$'s depending on the threshold. By choosing $F_1$ as evaluation metric, the threshold becomes another highly sensitive hyper-parameter of the model.

We also report on AUC (ROC) since it's more intuitive and independent of thresholds. An AUC of 0.5 represents no information or random predicitions which 1.0 is perfect information.

\section{System Architecture}
We have implemented four model architectures and the baseline is the logistic regression with two sets of features. The base copy of features are similar to \cite{DBLP:conf/emnlp/ClarkM16} namely:
\begin{itemize}
	\item Simplified version of sentence distance, including indicators for distances $<5$ as well as actual distance (rather than the 10 buckets found in the paper).
	\item Embeddings (from word2vec, 50 dimensions) of the first word, head words, last word, previous word, and next word.
	\item Speaker features, as described in the paper.
	\item Pair features, including whether or not there is an exact match, a relaxed head match or if there is the same speaker tagged.
\end{itemize}

\subsection{Model 1}
\begin{figure}[h]
	\begin{center}
		\includegraphics[width=0.8\linewidth]{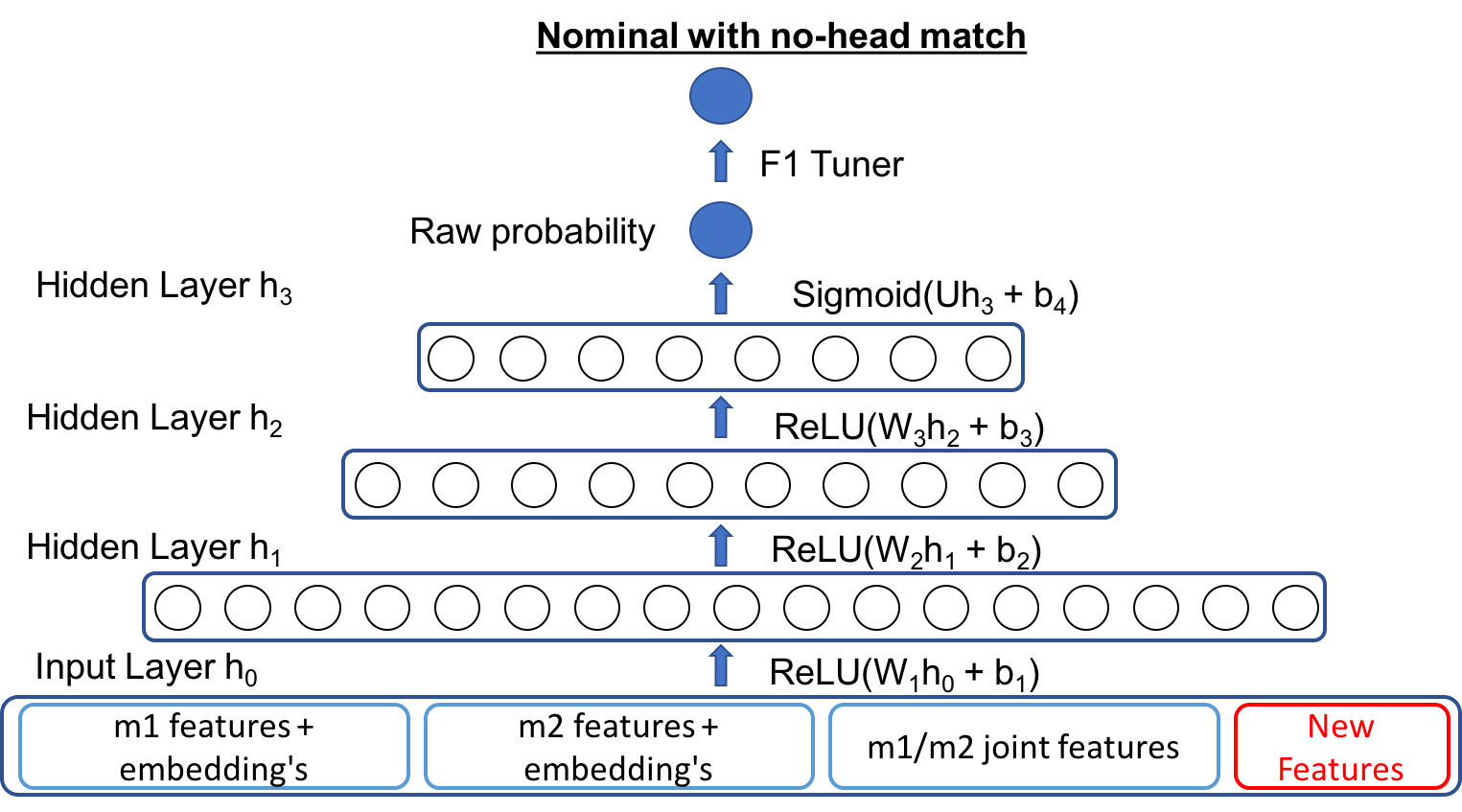}
	\end{center}
	\caption{Model 1 architecture with three layer feed-forward neural network}
	\label{model1}
\end{figure}

The starting point of our paper is the architecture proposed in \cite{DBLP:conf/emnlp/ClarkM16} (See figure \ref{model1}). The basic model is 3 fully connected layers with ReLU activations. The various hidden layer sizes are (500, 200, 100) and the final transformation is a sigmoid function. We use \textit{weighted cross entropy} as our loss function for training. We also experimented with \textit{L2 loss}, but found that weighted cross entropy better expressed the cost of missing rare events \-- a fundamental challenge in our problem. 

\subsection{Model 2}
\begin{figure}[h]
	\begin{center}
		\includegraphics[width=0.8\linewidth]{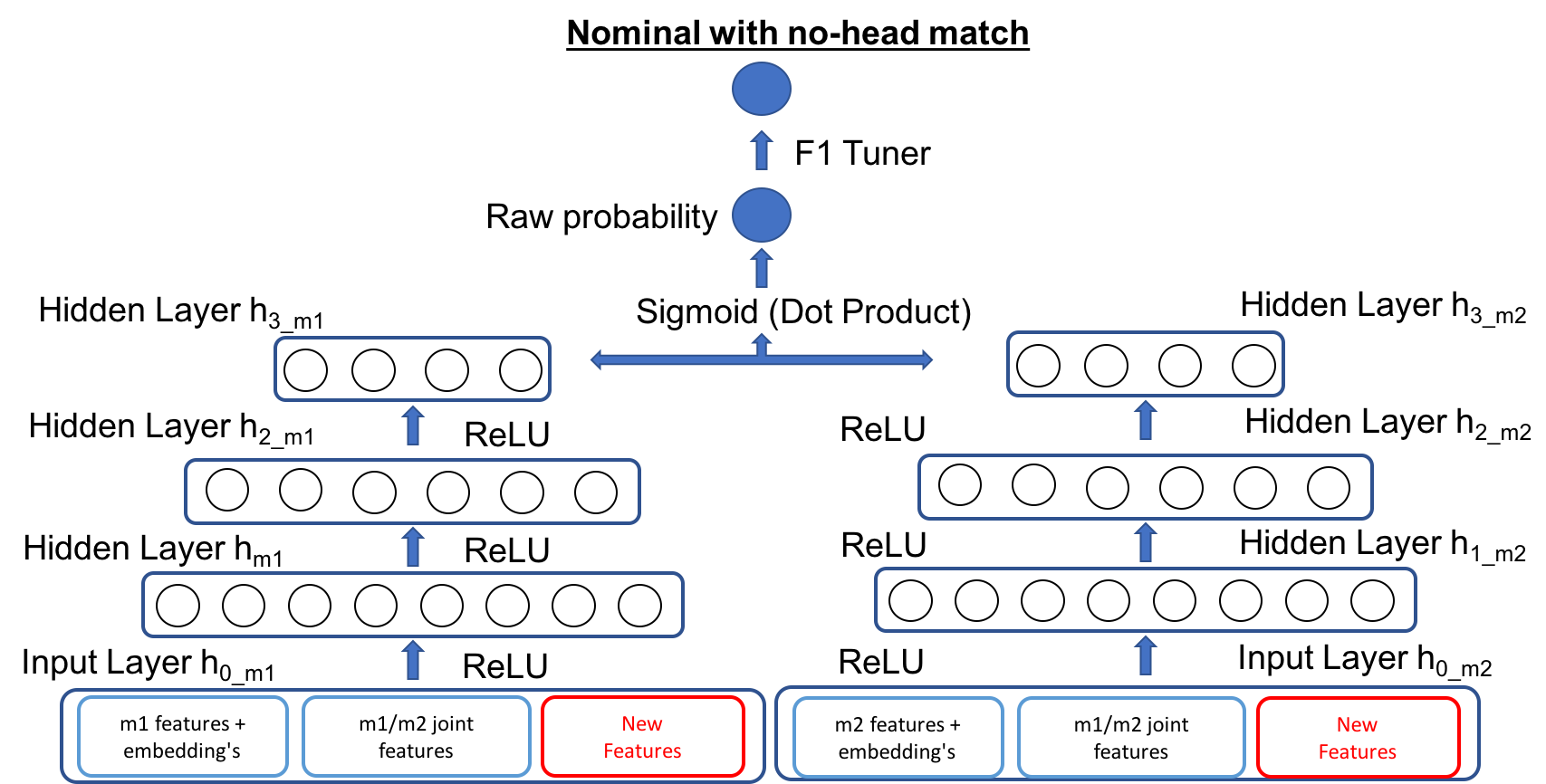}
	\end{center}
	\caption{Model 2 architecture learns the candidate antecedent and mention embeddings independently before combining them}
	\label{model2}
\end{figure}

A secondary structure focuses on taking the insight that we are essentially modeling some embedding interactions and creating 2 separate learning models, one for each mention. The input features are trimmed to remove features related to the other word (other than mention distance). The final hidden layer of each of these mention-level models are then combined in a final layer dot-product. We then run this final layer through the sigmoid function to come up with a final prediction. The 2 final layers that are inputs to the dot-product could be thought of as domain specific embeddings of each mention. The interaction of these embeddings are then translated to our probability of coreference. Refer to figure \ref{model2} for details.

\subsection{Composite Models (Model 3 \& Model 4)}
\textit{Model 3} is an ensemble of model 2 and model 3, where we simply add the 2 final scores from model 1 and model 2 and then apply a sigmoid function to that output. This model is also trained end-to-end rather than using pre-trained versions of model 1 and 2, otherwise they would end up over-predicting the outcome.

\textit{Model 4} is a modification onto model 3 where instead of simply adding the two models, we take the sigmoid function of each of model 1 and model 2. We then create a new submodel with the same set of input features to train a probabilistic model that puts a weight of \textit{prb} on model 1 and a weight of \textit{1-prb} on model 2. This probabilistic model is a logistic regression on input features that determines what weight should be given to model 1. This is also trained end-to-end. This allows the data to determine when model 1 may be preferable to model 2 and vice versa rather than do a simple ensemble average.

We explore regularization with \textit{dropout}, especially at the first hidden layer. We found that applying dropout at higher levels was detrimental to performance (see figure \ref{fig:sub2}). We believe that at the size of model we were testing, dropout ended up removing too many nodes to offset the benefit from generalizing.

\section{Experiments, Results \& Challenges}

\subsection{Results}
In our experiments we compare the baseline model of logistic regression with the four proposed models. The proposed models were tuned using the train and dev set on varying loss function, weights to the loss function, learning rate, neural network architecture, dropout and number of epochs. The best model performance on the dev set for each of the model was chosen and the results produced on the test set which are summarized in table \ref{tab:f1_scores_downsampled}. It must be noted that the test set, like the training and dev set has been downsampled to 2\% negative samples and uses the additional features as described in section \ref{subsec:addn_features}.
Model 4, training on 15 epochs with a slow learning rate of 0.0001 performs almost 12\% better than the baseline logistic regression while almost 8\% better than the tuned state-of-the-art model.

\begin{figure}
\centering
\begin{subfigure}{.5\textwidth}
  \centering
  \includegraphics[width=.95\linewidth]{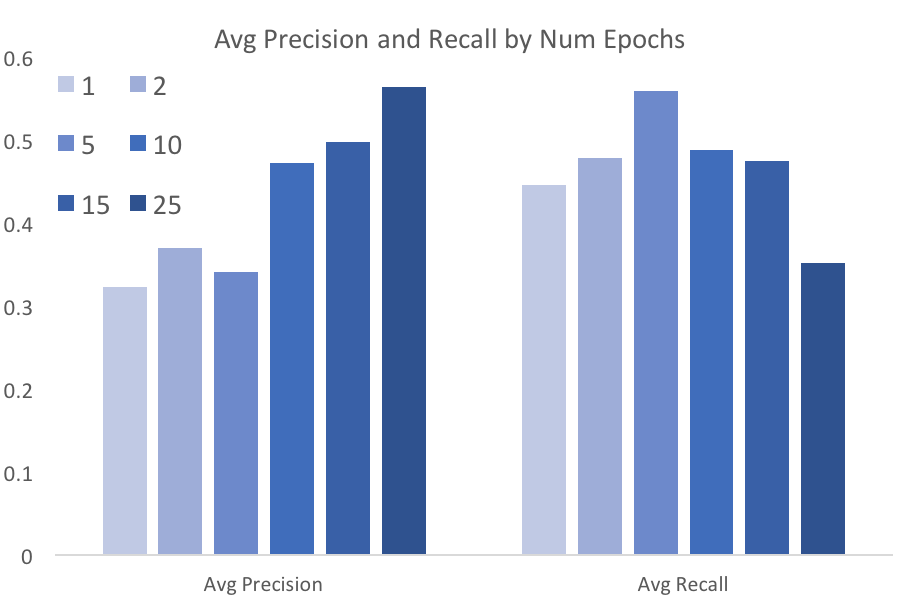}
  \caption{Average precision and recall for our models as \\ we increase the training epochs.}
  \label{fig:sub1}
\end{subfigure}%
\begin{subfigure}{.5\textwidth}
  \centering
  \includegraphics[width=.95\linewidth]{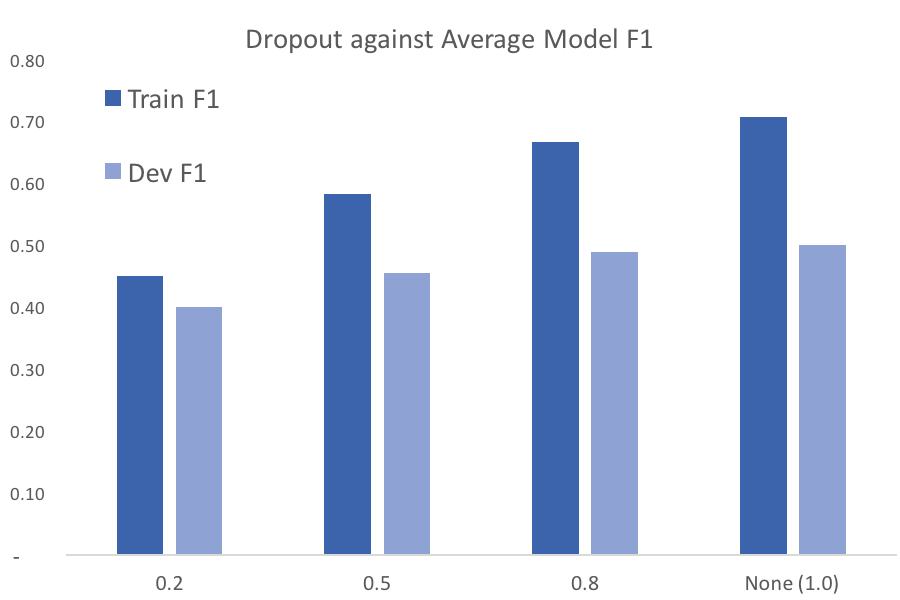}
  \caption{The train and dev $F_1$ scores with varying levels of dropout.}
  \label{fig:sub2}
\end{subfigure}
\end{figure}

Unsurprisingly, as can be seen in figure \ref{fig:sub1}, given the label imbalance, as we increase the number of epochs, we are able to increase the precision at the expense of recall for all models. The confusion matrix for the best performing model gives a good understanding of the labels and relative performance (Table: \ref{tab:model_4_confusion_matrix}).

\begin{table}[!htb]
	\begin{center}
		\begin{tabular}{l||c|c|c}
			\multicolumn{1}{c||}{\bf Model Name}  &\multicolumn{1}{c|}{\bf Training F1} &\multicolumn{1}{c|}{\bf Dev F1} &\multicolumn{1}{c}{\bf Test F1}

			\\ \hhline{====}
			Logistic Regression         & .76 & .36 & .49 \\
			Model 1             & .76 & \textbf{.53} & .53 \\
			Model 2             & .65 & .51 & .53 \\
			Model 3             & .75 & .52 & .56 \\
			Model 4             & \textbf{.82} & .52 & \textbf{.61} \\
		\end{tabular}
	\end{center}
	\caption{F1 score on 2\% downsampled negative examples data} 
	\label{tab:f1_scores_downsampled}
\end{table}

\begin{table}[!htb]
	\begin{center}
		\begin{tabular}{l||c|c}
			\multicolumn{1}{c||}{\bf}  &\multicolumn{1}{c|}{\bf Predicted NO} &\multicolumn{1}{c|}{\bf Predicted YES} 
			\\ \hhline{===}
			\textbf{Actual NO}         & 20272 & 424 \\
			\textbf{Actual YES}            & 559 & 531 \\
		\end{tabular}
	\end{center}
	\caption{Confusion matrix for Model 4 over test dataset.} 
	\label{tab:model_4_confusion_matrix}
\end{table}

\subsection{Impact of Additional Features}
\label{subsec:addn_features}
Looking at some of the errors like table \ref{tab:errors_with_addn_features}, especially false negatives for co-reference like "the molar-a tooth" and "my taxi-the car", we added the cosine distance of the word vector embeddings as raw features to the model. In particular, we added the dot-product between the first, last and the head word of the two mentions, creating four extra features for every mention pair based on the semantic relationship between the two mentions.
This drastically improved the model performance, as can be seen in table \ref{tab:f1_scores_addn_features}. Simply adding these features to the logistic regression increased the model performance by 4\%.
The intuition behind adding these features is simple \-- given concatenated word vectors $[m_1 m_2]$, a fully connected neural network can never learn $m_1' m_2$ which represents how similar $m_1$ is to $m_2$. Unsurprisingly, a lot of nominals with no head match are synonyms or specific example of a general entity which is now captured by these new features (this is also attempted by model 2).
\begin{table}[!htb]
	\begin{center}
		\begin{tabular}{l||c|c|c}
			\multicolumn{1}{c||}{\bf m1-m2 pairs}  &\multicolumn{1}{c|}{\bf P(Without Addn. Features)} &\multicolumn{1}{c|}{\bf P(With Addn. Features)} & \multicolumn{1}{c}{\bf $\Delta$}

			\\ \hhline{====}
			the public purse --- the public 's money & .0242 & .961 & +.93\\
			a molar – the tooth & .0043 & .740 & +.74\\
			the plunge - the downturn & .401 & .742 & +.341\\
		\end{tabular}
	\end{center}
	\caption{Example Wins: Examples of False Negatives without Addn Features turning to True Positive with Additional Features} 
	\label{tab:errors_with_addn_features}
\end{table}

\begin{table}[!htb]
	\begin{center}
		\begin{tabular}{l||c|c|c}
			\multicolumn{1}{c||}{\bf Model Name}  &\multicolumn{1}{c|}{\bf Training $F_1$} &\multicolumn{1}{c|}{\bf Test $F_1$} & \multicolumn{1}{c}{\bf Test AUC (ROC)}

			\\ \hhline{====}
			Without Addn. Features         & .71 & .45 & .90\\
			With Addn. Features            & .76 & .49 & .91\\
			\hhline{====}
			$\Delta$            & +.05 & +.04 & +.01\\
		\end{tabular}
	\end{center}
	\caption{The $F_1$ score improves drastically for logistic regression (actually, for all models) with the addition of the new features.} 
	\label{tab:f1_scores_addn_features}
\end{table}

\subsection{Challenges}
\textbf{Class imbalance}: As mentioned previously, the overall true rate of the examples in our dataset is approximately 0.1\%, meaning that models trained on the raw dataset have difficulty learning what distinguishes positive examples from negative examples. We take a multi-step approach to resolving this issue and guiding the model to learn to distinguish positive and negative examples.

We begin by downsampling our training data - keeping all positive examples and only 2\% of the negative examples. We also choose weighted cross entropy as our loss function, so that we can give further weight to positive examples through a tunable hyperparameter. When evaluating $F_1$ metrics, we are then forced to define a threshold at which we consider the model probability to be true through two mechanisms. 
\begin{itemize}
	\item When dealing with downsampled data, we use an $F_1$ tuner that searches over the training set for the threshold that yields the highest $F_1$, and apply this to our dev set evaluation.

	\item When training on downsampled data but evaluating on the full dataset (for our test set) we run our calibration process. This takes the distribution of predictions from the training set and finds the percentile of data such that the model will emit the correct true rate as though the data was not downsampled. For example, if the rate in our downsampled data is 33\% and the downsampling was 50\%, that means the true rate is around 17\%. In this example, calibration of the prediction percentile would be with the goal of yielding true 17\% of the time on the training set. The idea behind setting a higher threshold is that if we need to reduce the number of true predictions, we should only keep the highest conviction guesses from the model. In practice, the level of model certainty required to emit a True varied between 97.5\% to 99.8\% depending on the model.
\end{itemize}

\textbf{Memory and storage}: Because of the size of our dataset, we had challenges creating a full dataset on disk as well as fitting one in memory for training. Mitigating factors included only using half-precision for float and working with downsampled training data with a calibration layer. We also partitioned training data and running over all of the partitions each epoch. This greatly increased the time per epoch and experiment. A possibility to reduce the size of the data is to do in-memory lookups of word vectors, but this would have great slowed down the training speed as each example would require a read-and-process operation rather than a simple read. We could also use a databse to store the test/train and dev features.

\textbf{Evaluation and prediction reconciliation:} A full evaluation would involve creating a full set of CoNLL evaluation metrics on the entire dataset. Instead, we used $F_1$ on a simple pair-matching model without any reconciliation step.

\section {Conclusion and Future Work}
We found that extending the deep learning model by doing a inner product of the embeddings, a probabilistic model ensemble and adding the aforementioned set of features improves the model over the baseline. The first two improvements over a simple feed forward network gave our model more capacity and flexibility for identifying nominal co-references with no head match.

We believe there are several promising dimensions of future enhancements that we can explore along the following lines:
\begin{itemize}
\item \textbf{Downstream tasks}: Coreference links between mentions form an upstream part of a coreference system, and nominals with no head match are a fraction of the coreferences. In this paper, while we improve upon the existing state-of-the-art algorithm, we don't know the impact on the downstream tasks. 
\item \textbf{Document level memory}: We are looking at intra-document coreferences in this paper. Memory and attention play a role in human understanding of semantic information, which could be extended to this class of problems using LSTM or RNNs. It can also help weight multiword mentions to better capture the key part of what makes the object a real world reference.
\item \textbf{A separate layer for singleton mentions}: Identifying which mentions are never repeated could help the model by reducing the number of evaluated candidates for future evaluation. This could be done end-to-end or within a continuous framework with a penalty or probability interaction. This is motivated by \cite{conf/naacl/RecasensMP13} and could incorporate more semantic features.
\end{itemize}

\bibliography{cs224n_report_arxiv}

\end{document}